# Cognitive and Cultural Topology of Linguistic Categories: A Semantic-Pragmatic Metric Approach

**Eugene Yu Ji**
The University of Chicago
Chicago, IL, 60637
yuji1@uchicago.edu

## Abstract

In recent years, the field of NLP has seen growing interest in modeling both semantic and pragmatic dimensions. Despite this progress, two key challenges persist: firstly, the complex task of mapping and analyzing the interactions between semantic and pragmatic features; secondly, the insufficient incorporation of relevant insights from related disciplines outside NLP. Addressing these issues, this study introduces a novel geometric metric that utilizes word co-occurrence patterns. This metric maps two fundamental properties – semantic typicality (cognitive) and pragmatic salience (socio-cultural) – for basic-level categories within a two-dimensional hyperbolic space. Our evaluations reveal that this semantic-pragmatic metric produces mappings for basic-level categories that not only surpass traditional cognitive semantics benchmarks but also demonstrate significant socio-cultural relevance. This finding proposes that basic-level categories, traditionally viewed as semantics-driven cognitive constructs, should be examined through the lens of both semantic and pragmatic dimensions, highlighting their role as a cognitive-cultural interface. The broad contribution of this paper lies in the development of medium-sized, interpretable, and human-centric language embedding models, which can effectively blend semantic and pragmatic dimensions to elucidate both the cognitive and socio-cultural significance of linguistic categories.

## 1 Introduction

The field of NLP has recently witnessed a re-emerging interest towards incorporating pragmatics (i.e., meaning in context) and sociocultural meaning into computational language models (Hovy, 1990; Jurafsky, 2008; Kabbara, 2019; Shaikh et al., 2023; Li et al., 2020). Outside NLP, linguists have long recognized that pragmatics, which emphasizes language use and linguistic communication in context and often in culture, is inherently intertwined with semantics, the study of meaning itself (Hymes, 1974; Grice, 1975; Silverstein, 1993; Urban, 2006). The challenge for today's NLP researchers lies not just in understanding pragmatic properties, but in modeling between semantic and pragmatic knowledge, and cognition and culture in interpretable ways, a task that has proven non-trivial in the computational realm.

Over the past decade, NLP advancements, ranging from embedding techniques like word2vec and BERT (Mikolov et al., 2013; Devlin et al., 2018) to very impactful Large Language Models such as InstructGPT (Ouyang et al., 2022) and LLaMa (Touvron et al., 2023), have demonstrated remarkable prowess in modeling and generating semantic contents. However, the pragmatic language in cultural context and real-world language use, and its interrelation with semantics, is still relatively under-researched (Zellers et al., 2020; Min et al., 2022). The current NLP literature further accentuates this gap with occasional ad-hoc evaluation methods for pragmatics and culture and a limited assimilation of insights from related fields outside NLP.

In this paper, we present a new metric embedding technique to jointly embed semantic and pragmatic attributes within a hyperbolic space. Notably, we focus on basic-level categories, arguably one of most frequently investigated linguistic categories in both modern cognitive psychology and cognitive anthropology (Rosch, 1973, 1978; Rosch & Mervis, 1975; Jónsdóttir and Martin, 1996; Markman and Wisniewski, 1997; Hajibayova, 2013). On the other hand, while cognitive science research predominantly highlights the semantic dimension of basic-level categories, sociolinguistics and linguistic anthropology tend to emphasize their pragmatic and sociocultural facet (Silverstein 1993; Urban 2006). Bridging this gap, our paper offers an interpretable computational framework that integrates both dimensions and provides a quantitative exploration of their interrelationship. The paper proposes a "weighed-entropy" metric, and maps specific semantic and pragmatic properties in the low-dimensional hyperbolic space via "co-occurrence intensity" and "co-occurrence dominance" respectively derived from the metric.

## 2 Embedding semantic and pragmatic properties of linguistic categories

### 2.1 Semantics-driven embedding models in the literature

A substantial portion of embedding research in NLP is dedicated to interpreting results in terms of linguistic similarity, most often semantic categories. This encompasses various facets such as semantic associativity, semantic distance, the capability to represent similar or different world knowledge, etc. (e.g., Mikolov et al., 2013; Pennington et al. 2014). Such categorical semantic relationships are often depicted as vector relationships in a Euclidean space.

In particular, there has been a growing body of work that emphasizes deriving hierarchical linguistic relationships from word co-occurrence patterns. One primary example is hypernyms and hyponyms (Miller, 1995), such as in WordNet in English (Fellbaum, 1998). These hierarchical relationships can often be embedded as tree-like structures in Hyperbolic space, offering an alternative interpretable representation different from similarity representation. However, much less work in NLP has investigated modeling **semantic typicality**, an arguably more basic and prevalent dimension of semantic hierarchy than hypernym-hyponym relationship (Rosch & Mervis, 1975; Uyeda & Mandler, 1980, Banks & Connell, 2023; Plant et al., 2011), nor using it as a common computational baseline for evaluating NLP algorithms' performance on modeling semantic hierarchy. Mirsa et al. (2022) further demonstrate that semantic typicality may not be well captured by any current embedding or deep learning language models.

### 2.2 Cognitive and pragmatic/sociocultural interpretability of embedding models

In recent times, there has also been a surge in efforts to discern the psychological or cognitive and neurological relevance of word embeddings (Ettinger et al., 2016; Toneva & Wehbe, 2020; Goldstein et al., 2022; Caucheteux et al., 2022). These endeavors aim to bridge the gap between computational models and human cognition and provide a powerful lens for interpreting embedding models and their results.

Despite the relatively extensive research on semantic and cognitive interpretability, there has been a remarkable lack when it comes to pragmatic interpretability. This underscores a critical area of potential exploration and innovation in the domain of NLP embeddings, which has significant implications for pragmatic language models beyond semantic embedding ones.

Inspired by work in philosophy of language, sociolinguistics, and linguistic anthropology (Putnam, 1975; Silverstein, 1976, 1993; Urban, 2006), this work attempts to incorporate **pragmatic salience**, which measure how much a word or linguistic expression is pragmatically and socioculturally salient given a native speaker's pragmatic repertoire, as another fundamental language embedding dimension (for example, intuitively, the term "white" in American English has a stronger and more salient pragmatic and cultural connotation about race than "orange"). Importantly, this line of work also emphasizes the intrinsic relationship between the pragmatic repertoire of natural languages and sociocultural relationships[1].

### 2.3 Developing a cognitive-sociocultural embedding model through basic-level categories

Basic-level categories, though not extensively explored computationally in Natural Language Processing (NLP), have been a focal point of study in cognitive science, sociolinguistics, and linguistic anthropology. Through the lens of human cognition and semantic structures, cognitive science and psychology examine semantic categories that are categorically neither overly broad nor excessively specific, with seminal work by Brown (1958), Rosch (1973, 1978), Rosch & Mervis (1975), and more recent computational analyses by Regier et al. (2015, 2017). These studies delve into how basic-level categories (such as “red” and “blue” rather than their hypernyms like “color” or their hyponyms like “reddish” or “azure”) facilitate early language acquisition and semantic categorization, despite ongoing debates regarding their developmental primacy and cognitive processing (Mandler, 1988, 1993; Rogers & Patterson, 2007). In contrast, sociolinguistics and linguistic anthropology view basic-level categories from a sociocultural perspective, emphasizing their formation and use within specific cultural and social contexts (Sahlins, 1976; Lucy, 1997; Silverstein, 2022). This divergence underscores the intricate interplay between cognitive, semantic, and sociocultural factors in shaping linguistic categories, highlighting the need for a multidisciplinary computational approach.

Though NLP community began to explore basic-level categories in recent years (Mills et al. 2018; Renner et al. 2023; Misra et al. 2021), they remain under explored. Several challenges contribute to this gap: (1). Definition Complexity: Basic-level categories, as first defined in the literature of cognitive science, present a multifaceted and intricate landscape (Berlin & Kay, 1969; Rosch, 1973, 1978; Rosch & Mervis, 1975; Barsalou, 1983; Mahon & Caramazza, 2009; Pothos & Wills, 2011). The current absence of automated extraction methods of basic-level categories from textual repositories, such as WordNet (Fellbaum, 1998), further exacerbates the challenge of integrating these categories into small-scale embedding models. (2). Specificity Constraints: Basic-level categories are inherently domain-specific and often language-bound. Given the lack of robust and interpretable automated extraction techniques, incorporating nuanced domain-centric and language-specific attributes into the training and evaluation phases of embedding models becomes even more difficult.

## 3 A Semantic-pragmatic metric approach

### 3.1 The weighted-entropy metric

The model intends to offer a new metric-driven method for measuring and mapping both basic semantic and pragmatic properties through word co-occurrence statistics. Specifically, the first step is to derive the metric through “weighted-entropy”, which is defined as follows: given the target word ’s conditional entropy upon the raw co-occurrence statistics of its context words, it is weighted by two parameters: the first is the target word’s collocation diversity and the second is the ranking of the conditional entropy of each pair of word occurrences for every target word. While the first parameter facilities extracting categorical relationships from corpus, the second parameter endows the metric with a hierarchical structure.

Theoretically, proposing collocation diversity and entropy ranking as two important weighing parameters is motivated by complex network theory. The concept of collocation has been widely used in existing work of semantic modeling to characterize linguistic similarity and associativity relationship (Sinclair, 1991; Manning & Schütze, 1999; Evert, 2009; Vincze, 2015). Probabilistic ranking, on the other hand, is demonstrated to be related to the hierarchical characteristics of complex networks as well the curvature of hyperbolic space which can embed network structures (Ravasz & Barabási, 2003; Krioukov et al., 2010; Papadopoulos et al., 2012). Hyperbolic word embedding has grown interests in recent NLP literature (Nickel & Kiela, 2017; Sala et al., 2018; Dhingra et al., 2018; Tifrea et al. 2019).

[1] In this paper, the term “pragmatic salience” refers to a word’s degree of pragmatic explicitness in contrast with the overall pragmatic knowledge and repertoire that a native speaker possesses, rather than modeling specific pragmatic usage in communication, on which NLP researchers have extensively investigated (Frank & Goodman 2012; Benjamin et al. 2023; Jurafsky, 2008; Kabbara, 2019; Shaikh et al., 2023; Li et al., 2020).

In cognitive science, research has also argued that collocation can be related to language processing in the brain such as non-frequency-based syntactic and semantic knowledge, cognitive attention, and priming (Vespignani et al., 2010; Arnon & Snider, 2010; Durrant & Doherty, 2012). More recently, computational neuroscientists and cognitive psychologists also use probabilistic ranking and hyperbolic geometry to model categorization in sensory, perceptual, and cognitive processing (Ganea et al., 2018; Sharpee, 2019; Sengupta et al., 2019). This paper extends these two cognitively meaningful measures as two weighting parameters in constructing the new linguistic metric.

Formally, let $w_i$ be a target word and $w_j$ be any context word, both selected from the whole vocabulary $W$ of the corpus. For a given target word $w_{i=a}$ , the measure of collocation diversity, denoted as $d$, is defined as the ratio $count\ (w_j)/f(w_a)$, where $f(w_a)$ represents the frequency of the target $w_a$ within the corpus or language input data, and $count\ (w_j)$ refers to the count of distinct context words $w_j$ that co-occur within a specific context window adjacent to $w_a$ . Additionally, the parameter $r$ is determined by the rank of conditional probability $f(w_{i=a}|w_{j=b})$ among all probabilities $f(w_{i=a}|w_j \in W)$, where $w_{i=a}$ is the target word, $w_j$ is the context word of $w_a$, and $W$ represents the entire vocabulary. The parameters $d$ and $r$ are conveniently combined into a single weighting parameter $q$, as illustrated in Equation 1.

The weighted entropy-based metric $H$ for embedding is defined as follows:

$$H(w_{j=b}) = -q_{ab} f(w_a, w_b)\ log(q_{ab} \frac{f(w_a, w_b)}{f(w_b)})$$

$$= -q_{ab} f(w_a, w_b)\ log(q_{ab} f(w_a|w_b)\ ) \qquad (1)$$

$$\text{with} \quad q_{ab} = d_{i=a} \frac{1}{(r_{i=a,j=b}+1)}$$

$$= \frac{count\ (w_j)}{f(w_a)} \cdot \frac{1}{1+rank[f(w_{i=a}|w_{j=b})\ in\ f(w_{i=a}|w_j \in W)]} \qquad (2)$$

Equation (1) represents the formula for the weighted conditional-entropy metric $H$, which quantifies the weighted co-occurrence relationship between the target word $w_a$ and any context word $w_b$ , given the textual input. Equation (2) details the formula of the weighting parameter $q_{ab}$, which is a linear function of $d$ and $r$. Both $w_a$ and $w_b$ are extracted from the vocabulary of the whole corpus of language input $W$.

The model defines a distributional metric $C$ for each target word $w_{i=a}$ within the range of [0,1], as the expected value of the sigmoid function of $H$. Let $S$ be the sigmoid function of $H$, for a given target word $w_{i=a}$ , $C(w_{i=a})$ can be expressed as followings:

$$C(w_{i=a}) = < S(H(w_{i=a}|w_j)) >$$

$$= < \frac{1}{1+exp\ (-H(w_{i=a}|w_j)} > \qquad (3)$$

Here, $C(w_i)$ serves as the metric for extracting the semantic and pragmatic dimensions for any target word $w_i$ . The first weighting function collocation diversity $d$ enhances the ability to abstract linguistic categories at a high level (i.e., the basic level), while the second weighting function probabilistic ranking $r$ endows the metric with hierarchical properties. The complete matrix $M^C$ formed by elements $C(w_i|w_j)$ is calculated with a specific minimum frequency threshold and a specific context window size.

### 3.2 Generating the mapping of basic-level categories by the weighted-entropy metric

Typical dimensional reduction algorithms often yield low dimensional yet not usually interpretable representations. To avoid this interpretability problem, generating the low-dimensional mapping of basic-level categories under the weighted-entropy metric is performed via a simple procedure of generating word sets as linguistic categories. Let $l(w_i)$ represent the count of non-zero values of $C(w_i|w_j)$ for $w_i$ . For a chosen target word $w_a$, let $l^{[w_a]}(w_i)$ denote the count of non-zero values of $C(w_i|w_j)$ and $C(w_a|w_j)$ for every context word $w_j$ shared by target word $w_a$ and any target word $w_i$ from the entire vocabulary $W$. The strategy for extracting $W_i^{[w_a]}$, a set of words linguistically closest to $w_a$, is outlined in Equation 4:

$$W_i^{[w_a]} = \left\{ w_{i\ max}^{[w_a]} \right\}$$

$$\text{where } w_{i\ max}^{[w_a]} = argmax_{w_i}(L^{[w_a]}(w_i))$$

$$\text{with } L^{[w_a]}(w_i) = \frac{2\ log\ l^{[w_a]}(w_i)}{log\ l(w_i) + log\ l(w_a)} \qquad (4)$$

The word set $W_i^{[w_a]}$ produced through the procedure in Equation (4), comprises words deemed categorically closest to $w_a$, with the distance between each $w_i$ and $w_a$ determined by $L^{[w_a]}(w_i)$. This procedure nominates candidate basic-level categories as word sets, while preserving the low-dimensional characteristic of the metric C.

### 3.3 Embedding semantic hierarchy and pragmatic salience through "co-occurrence intensity" and "co-occurrence dominance" in the hyperbolic space

Target words elected by the procedure characterized by Equation 4 are mapped into a two-dimensional Euclidean space. For any target word $w_{i=a}$, let

$$x = C(w_{i=a}) = < S(H(w_{i=a}|w_j)) >,$$
$$y = C(w_j) = < S(H(w_j|w_{i=a})) > . \quad (5)$$

where the x-axis is the expected value of the sigmoid function of the target word $w_{i=a}$'s weighted conditional entropy, and the y-axis, on the other hand, is the expected value of the sigmoid function of context words $w_j$'s weighted entropy conditional on the target word $w_i$. Euclidean coordinates (*x, y*) are then transformed into hyperbolic coordinates (*u, v*):

$$u = In\sqrt{\frac{x}{y}}, v = \sqrt{xy}. \quad (6)$$

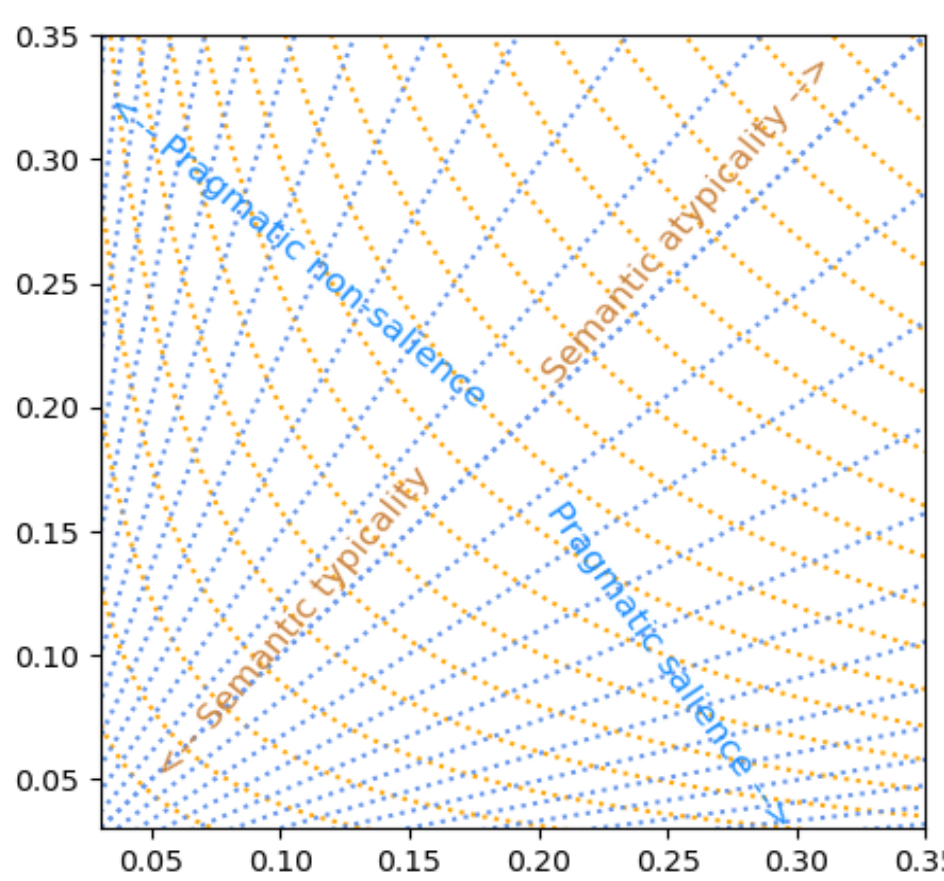

Figure 1. Illustration of the embedding coordinates that semantic-pragmatic approach generates. The *u* coordinate (in blue) represents pragmatic salience and non-salience, and the *v* coordinate (in brown) denotes semantic typicality and atypicality (semantic hierarchy) for basic-level categories.

In this approach, the *u* coordinate is introduced as a basic measure of pragmatic salience and the *v* coordinate as a basic measure of semantic typicality (as semantic hierarchy) for basic-level categories (Figure 1). Intuitively, the *u* coordinate quantifies the *co-occurrence intensity* between word $w_{i=a}$ and its context $w_j$, gauging how frequently one word co-occur with its overall context (determined by both how $w_{i=a}$ serves as a target word with $w_j$ s as its context and as a contextual word of $w_j$ s), measured together with other $w_i$ s by the same weighted-entropy metric. On the other hand, the *v* coordinate captures the *co-occurrence dominance* between $w_{i=a}$ and any $w_j$, assessing the extent to which one word influences the co-occurrence when it acts as the target word compared to when it is a contextual word.

Drawing from concepts in sociolinguistics and linguistic anthropology, the *v* coordinate can be understood as indicating the degree to which a word shapes the genre or register of a text relative to its surrounding words (Silverstein, 1993; Urban, 2006). On the other hand, the representation of semantic hierarchy in hyperbolic space, often through Poincaré disk models, has gained traction in recent NLP research (Nickel & Kiela, 2017; Sala et al., 2018; Dhingra et al., 2018; Tifrea et al. 2019). Yet, such modeling predominantly focuses on a thinly represented hypernym-hyponym semantic hierarchy. The semantic-pragmatic (*u, v*) coordinates present a framework that not only captures a potentially more foundational semantic hierarchical dimension rooted in cognitive science but also integrates a pragmatic hierarchical relationship. Furthermore, the hyperbolic coordinates employed here are compatible with the metric space structure of the Poincaré half-plane model of hyperbolic geometry. This compatibility ensures that while computationally simple as they are, the (*u, v*) coordinates can be aligned with coordinate systems used in other linguistically thinner yet computationally tractable hyperbolic embedding models as mentioned above.

## 4 Results and evaluations

### 4.1 Extracting domain-specific basic-level categories

The first experiment utilizes the procedure specified in Section 3.2 to extract basic-level categories from the corpus data. The data contains two corpus of English Wikipedia texts randomly chosen from the whole English Wikipedia data. Each of the corpus contains 8M word tokens. Experiments are performed with a range of minimum target word frequency threshold (5, 10, 15, 20, 30, 50) and context window size (1, 2, 3, 5, 6, 10). Testing shows no significant discrepancies in terms of the basic-level categories being selected by the algorithm, once minimum target word frequency threshold is greater than 15 and smaller than 50, and context window size is greater than 5 but smaller than 10. In actual mapping, 15 and 5 are selected as the frequency threshold and context window size respectively for optimal effectiveness and efficiency. Further testing also shows that for most selected basic-level categories, training begin to converge around 5 to 6 M tokens and increasing the size of the training Wikipedia corpus to 20 M does not impact the results overall.

Our empirical analysis has been conducted on 34 semantic domains (mostly formed by nouns and adjectives), which are selected primarily according to literature investigating semantic typicality (as basic semantic hierarchy) in cognitive psychology in cognitive psychology (Rosch & Mervis 1975, Uyeda & Mandler 1980, or Banks & Connell 2023). Among the 34 investigated linguistic domains, 680 words are extracted by the weighted-entropy metric as linguistically closest to a target word in each domain (See sections 3.2 and 3.4), among which 213 words (31.3%) can be identified as basic-level categories, which means that any of them appears at least once as words under the same domain elicited by experiments on human cognitive typicality ranking, as semantic hierarchical relationship, in at least one literature on cognitive semantic typicality listed above. For each semantic domain, 3 words with relatively high typicality are chosen as target words. Besides the 3 words with highest typicality, if any additional word among the 20 words closest to $w_a$ , as determined by Equation 4, also appears in any of the word lists in the cognitive studies mentioned above, then the word will also be selected as a basic-level category belonging to the set $W_i^{[w_a]}$.

### 4.2 Mapping semantic hierarchy and pragmatic salience in hyperbolic coordinates

The 34 domains are further identified manually the following types: (a). Domains that have been extensively investigated in cognitive psychology and anthropological linguistics for both their semantic and pragmatic properties, examples of which include basic-level categories of color (Figure 2a; Berlin & Kay, 1969; Sahlins, 1976; Lucy, 1997; Zaslavsky et al., 2019; Silverstein, 2022) and kinship (Figure 2b – 2d; Hirschfeld, 1986; Kemp & Regier, 2012). (b). Domains that have clear pragmatic and sociocultural significance, yet the relationship between their semantic and pragmatic properties (Figure 3a and 3b) have not been extensively studied in cognitive science or NLP. Examples of such a type include political identity and religious identity. (c). Domains that have been richly investigated for their semantic attributes in NLP embedding, but often have their pragmatic aspects overlooked (Mikolov et al., 2013; Bennington et al., 2014). Due to the paper's length limit, the rest of this section mainly focuses on an in-depth analysis of a few domains respectively selected from the first two types.

| **Models** | **Basic-level color categories in English** |
|---|---|
| Cognitive studies (Rosch & Mervis 1975; Regier et al. 2015, 2017) | Black, white, red, yellow, green, blue, orange, purple, brown, purple, pink, gray |
| Corpus-driven embedding (w*ord2vec* from Mikolov et al. 2013) | Black, white, red, yellow, green, blue, orange, purple, brown, purple, pink |
| Corpus-driven embedding through the weighted- entropy metric of this paper | Black, white, red, yellow, green, blue, orange, purple, brown, purple, pink, gray |

Table 1: Comparisons between empirical studies on basic-level color categories in English and results from embedding models. For extracting basic-color categories from corpus data, the weighted-entropy metric proposed in this paper performs slightly better than the classical *word2vec* embedding algorithm, with word2vec missing one color term "grey". Data size = 8M for both algorithms.

For extracting basic-level categories from corpus data, the weighed-entropy metric proposed in section 3.2 performs similarly overall as typical semantic embedding models such as *word2vec* (See Table 1 for a comparison of results on basic-level color categories). But most cognitive or embedding models in the literature do not map culture-driven pragmatic knowledge and their relationships with semantic one. Instead, equations (5) derived in Section 3.3 map words that are selected or extracted from the previous procedures to the two-dimensional Euclidean space and then to the hyperbolic coordinates ($u$, $v$), where a geometrical mapping for both semantic and pragmatic properties emerge, with the $v$ coordinate representing semantic hierarchy (typicality and atypicality) and the $u$ coordinate representing pragmatic salience and non-salience.

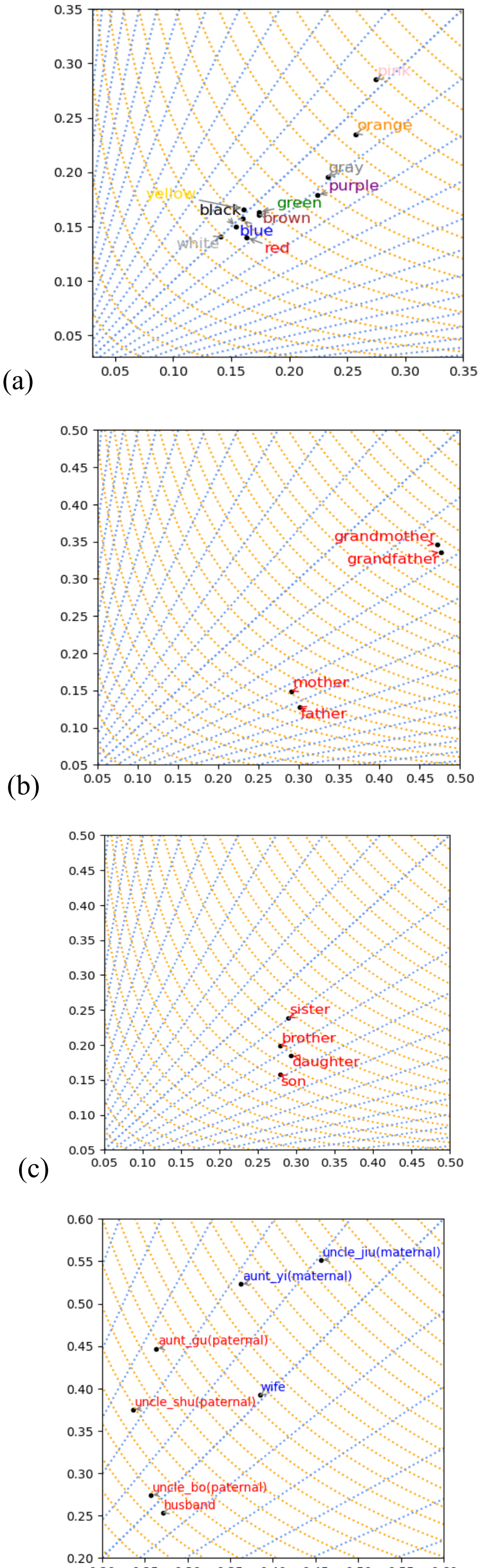

Figure 2. (a) – (c): The semantic-pragmatic mapping of basic-level color and selected kinship categories in American English. (d): In contrast, the semantic-pragmatic mapping of selected paternal and maternal kinship categories in standardized Chinese shows a patrilineal kinship order.

Figure 2 (a) to (c) presents the outcomes for basic-level color and kinship categories. The color terms (Figure 2a) generally adhere to the cross-cultural color-term semantic typicality rankings found in cognitive science literature (Table 1). The kinship terms that the model captures in general resemble those extracted from other classical word embedding models (e.g., Mikolov et al., 2013; Pennington et al., 2014). But it offers an additional dimension of interpretation other than then semantic similarity one. In particular, results in English show that both "father" ($u$ = 0.8551, $v$ = 0.1963) and "mother" ($u$ = 0.6694, $v$ = 0.2082) show higher pragmatic salience (higher $u$ values) and semantic typicality (lower $v$ values) than "grandfather" ($u$ = 0.3504, $v$ = 0.4003) and "grandmother" ($u$ = 0.3105, $v$ = 0.4041) (Figure 2b). In comparison, "son" ($u$ = 0.5686) and "daughter" ($u$ = 0.4598) show consistently higher pragmatic salience than the "brother" ($u$ = 0.3379) and "sister" ($u$ = 0.3105), but their values of semantic typicality are much closer to one another (with each $v$ value = 0.2100, 0.2328, 0.2356, 0.2633 respectively) (Figure 2c). In contrast, under the same training parameters and a randomly selected Chinese Wikipedia corpus of the same size, the models identifies a patrilinear kinship order among the basic-level kinship terminologies in standardized Chinese (Figure 2d), in which paternal terms (in red) are mapped as having both higher semantic typicality overall (towards bottom left measured by the brown dotted hyperbolic lines in Figure 2d) and higher pragmatic salience overall (towards bottom right measured by the blue dotted lines in Figure 2d) than maternal terms (in blue).

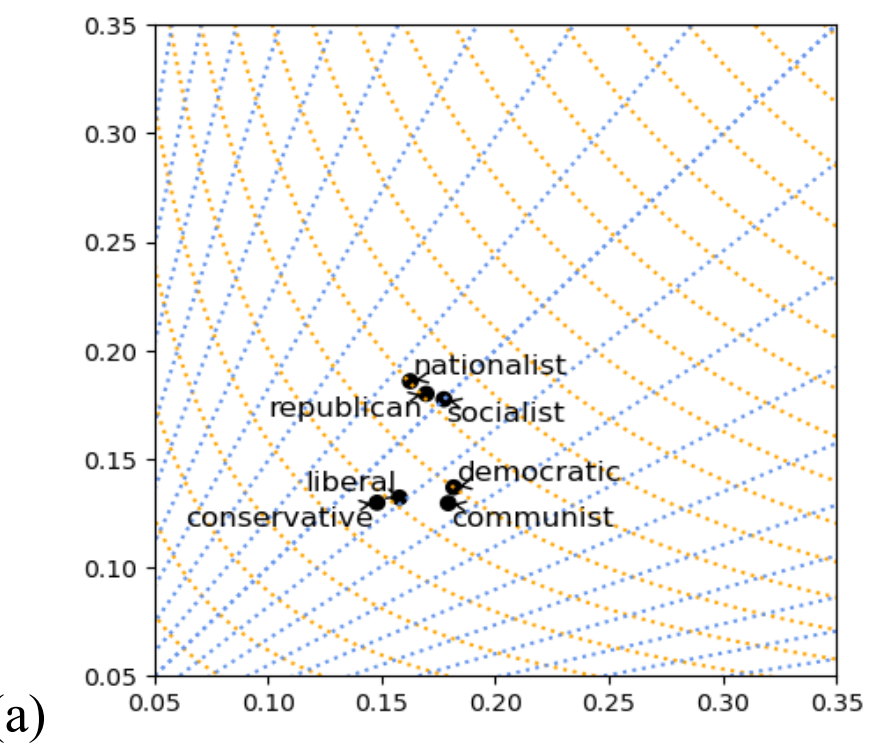

(a)

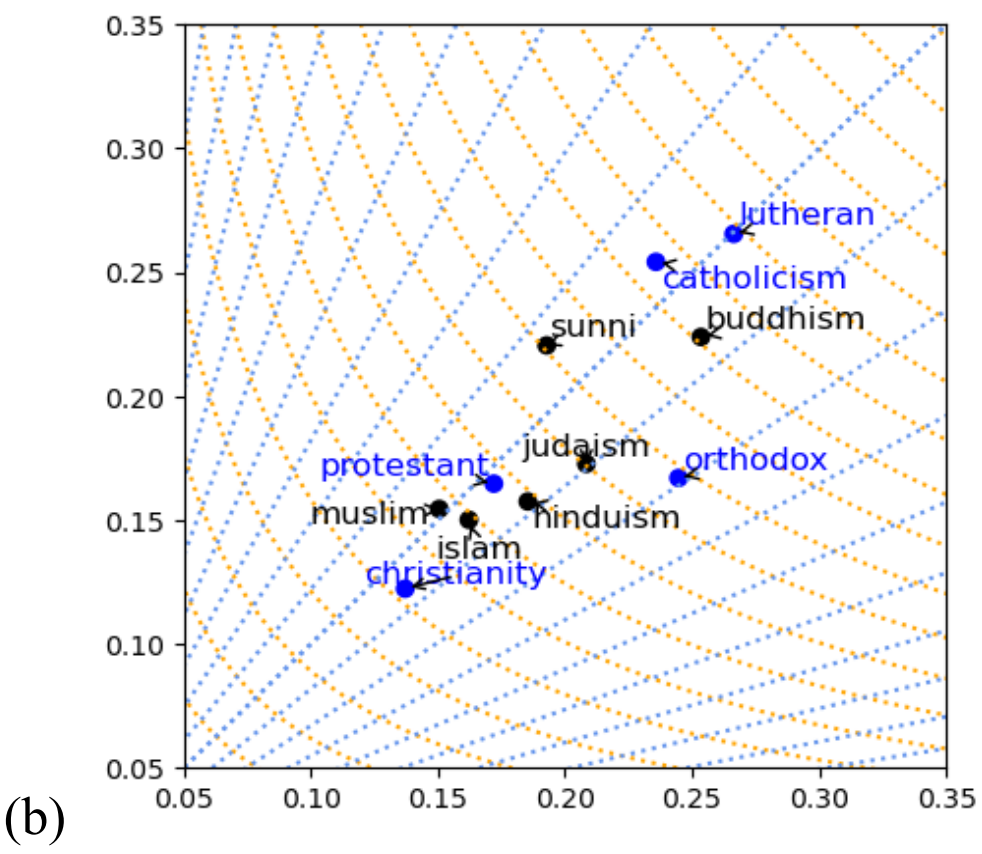

(b)

Figure 3: The semantic-pragmatic mapping of basic-level political and religious identities in American English. (3a) "Conservative" and "liberal" are semantically typical, while "democratic" and "communist" show high pragmatic salience, arguably reflecting Cold War ideological tensions. (3b) "Christianity" is most typical, with non-Christian religions overall surpassing specific Christian denominations in typicality. "Orthodox" displays the highest pragmatic salience, suggesting more pronounced intra-Christian varieties (especially with "Catholicism") than those between "Christianity" and other major non-Christian religions.

Applied to map political and religious identities based on written American English Wikipedia corpus (size = 8M), the results are also meaningful. For basic words of political identities (adjectives), identities such as "conservative" and "liberal" emerge as the most semantically typical (with lowest $v$ values 0.1386 and 0.1444 respectively). On the other hand, "democratic" and "communist" exhibit the highest pragmatic salience (with highest $u$ values 0.2813 and 0.3246 respectively). This result arguably echoes the pragmatic, historical context of the Cold War era, for which democracy and communism have often been projected as the most prominent pair of ideological competitors (Figure 3a; Fukuyama, 1989; Gaddis, 2005; LaFeber, 2008). Regarding religious identities, the generic term "Christianity" is mapped as the most typical. Major non-Christian religions, such as "Islam", "Muslim", "Judaism", "Sunni", and "Buddhism", exhibit higher typicality than specific Christian denominations like "Protestant", "Orthodox", "Catholicism", and "Lutheran". From a pragmatic perspective, "Orthodox" displays the highest salience among all religions, possibly reflecting religious tensions between Orthodox Christianity and other Christian denominations. Furthermore, the model suggests that the pragmatic distinction between "Orthodox" and "Catholicism" is more pronounced than that between "Orthodox" and "Protestant" or between "Orthodox" and "Lutheran", a finding that resonates with the deep, cultural-historically entrenched tensions between Orthodox Christianity and Catholicism that have persisted since the Great Schism in 1054 (Meyendorff, 1995; Ware, 1997; Papanikolaou & Demacopoulos, 2013).

| **Domains** | **SD on the dimension of pragmatic salience (*u*)** | **SD on the dimension of semantic typicality (*v*)** |
|---|---|---|
| Color | 0.0850 | 0.0444 |
| Kinship | 0.1979 | 0.0700 |
| Political identities | 0.1736 | 0.0157 |
| Religious identities | 0.1419 | 0.0430 |

Table 2. Standard Deviations on the dimension of pragmatic salience and semantic typicality for four domains of linguistic categories in modern English.

Together, these results show the promise of the semantic-pragmatic metric approach in mapping the interplay between semantic and pragmatic properties of linguistic categories within and across domains and languages (Table 2).

## 5 Conclusion

This paper presented a semantic-pragmatic metric approach, a novel approach that co-embeds semantic typicality and pragmatic salience of basic-level categories within a 2D hyperbolic space. Drawing upon interdisciplinary insights from cognitive science to linguistic anthropology, this model aims to provide a highly effective and interpretable computational model investigating the interplay between semantic and pragmatic dimensions and the interface between cognitive and cultural significance of linguistic categories.

## Limitations

This study, while pioneering in its approach, acknowledges several limitations. Firstly, the pragmatic applicability of all basic-level domains is not guaranteed, as the computational quantification of pragmatic meaning remains an intricate challenge. This complexity underscores the necessity for further engagement with empirical studies in disciplines such as pragmatics, sociolinguistics, and linguistic anthropology to establish more robust and comprehensive quantitative baselines for assessing pragmatic salience. Additionally, the advent of Language Learning Models (LLMs) incorporating human feedback, as seen in works like Ouyang (2022) and Touvron et al. (2023), also presents a promising avenue for the development of interface-based empirical baselines in pragmatic modeling.

Secondly, while multilingual testing is currently in progress with encouraging preliminary results, comprehensive analysis is required to ascertain the efficacy of the semantic-pragmatic approach in multilingual scenarios. This involves an extensive evaluation to understand the model's adaptability and performance nuances for basic-level categories across languages.

Thirdly, the scope of the study warrants expansion beyond the Wikipedia database to encompass various genres of written texts, particularly those from informal contexts and spoken language. The wealth of large-scale online informal texts and spoken language transcripts now accessible online, such as social media conversations, customer reviews, forum discussions, and podcast transcripts, present invaluable resources. Modern modeling efforts addressing the categorical knowledge embedded in natural language must integrate these diverse and nuanced communication forms to develop a more holistic and diverse understanding of both linguistic pragmatics and semantics.

Lastly, especially in many non-Latin-script languages, the scope of basic-level categories often extends beyond single-word or single-character entities. Although the semantic-pragmatic model is theoretically equipped to accommodate multi-word or multi-character tokens, this capacity necessitates further exploration and validation through rigorous testing.